\documentclass[letterpaper, 10 pt, conference]{ieeeconf}  

\IEEEoverridecommandlockouts                              

\usepackage{graphicx}
\usepackage{caption} 
\usepackage{hyperref} 
\usepackage{amsmath}  
\usepackage{amssymb}  
\usepackage{bm}       
\usepackage{booktabs}

\hypersetup{
colorlinks=true,
linkcolor=black
}

\title{\LARGE \bf
PMG: Parameterized Motion Generator for Human-like Locomotion Control
}

\author{
Chenxi Han$^{1,2,*}$,
Yuheng Min$^{1,2,*}$,
Zihao Huang$^{2}$,
Ao Hong$^{1}$,
Hang Liu$^{3}$,
Yi Cheng$^{2,\dagger}$,
Houde Liu$^{1,\dagger}$\\
$^{1}$ Tsinghua University \quad
$^{2}$ ZERITH Robotics  \quad
$^{3}$ University of Michigan \\
$^{*}$ Equal contribution \quad
$^{\dagger}$ Corresponding author
}

\begin{document}
\let\oldtwocolumn\twocolumn
\renewcommand\twocolumn[1][]{%
    \oldtwocolumn[{#1}{
    \begin{flushleft}
        \centering
        \vspace{-30pt}
        \includegraphics[clip,trim=0cm 0cm 0cm 0cm,width=0.92\textwidth]{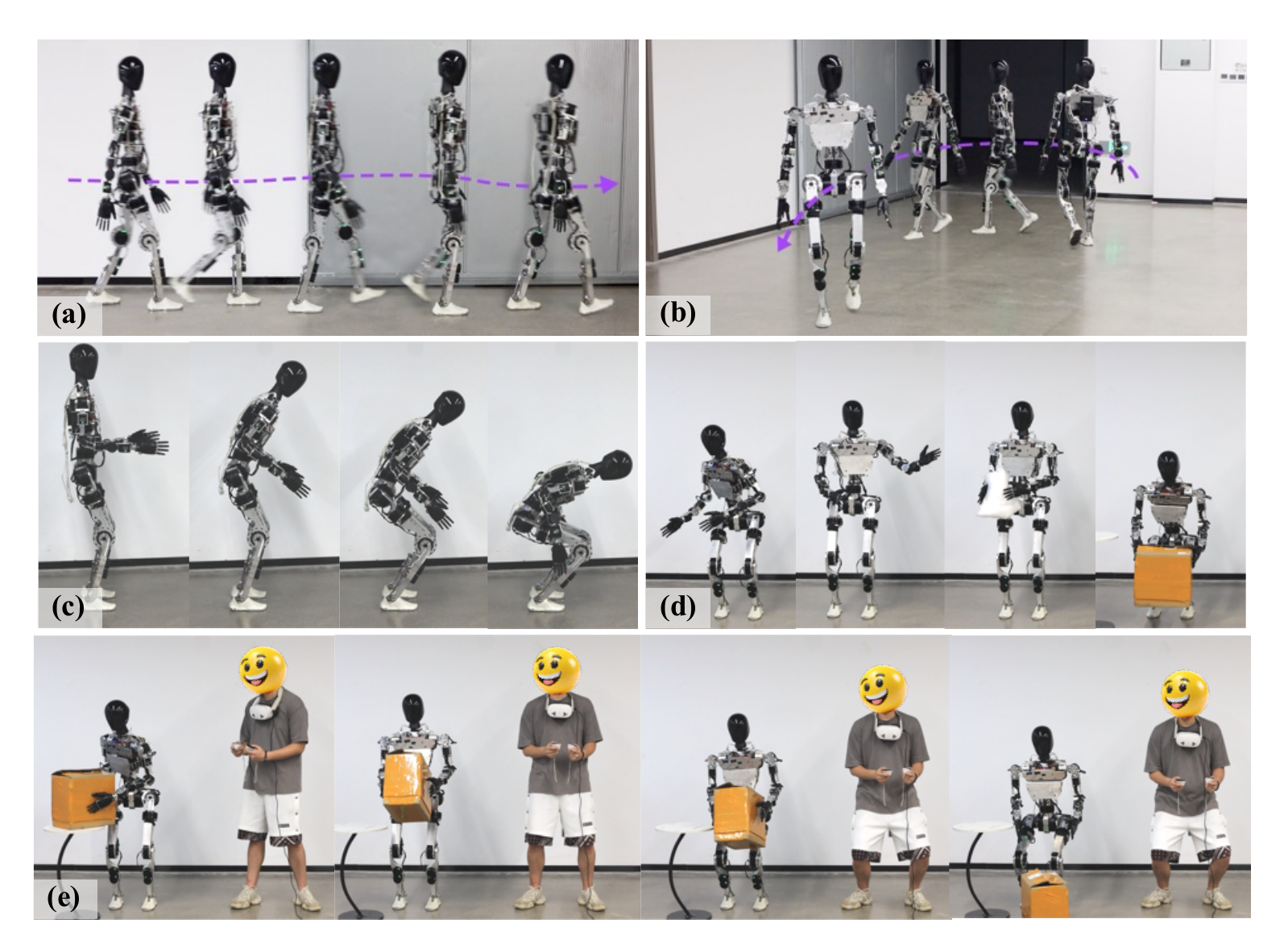}
        \captionsetup{justification=justified} 
        \captionof{figure}{
        Real-robot demonstrations of the proposed PMG on the ZERITH Z1 platform. 
        (a) Human-like walking gait generated beyond the reference motion dataset. 
        (b) Flexible omnidirectional walking achieved using less than 10 seconds of parameterized dynamic data. 
        (c) Wide-range, natural squatting and bending motions generated from only two clips of parameterized static data. 
        (d) Precise full-body pose and teleoperation control enabled by PMG. 
        (e) Teleoperated manipulation task of placing a box on the ground, demonstrating generalizability to diverse manipulation tasks.
        }\label{fig:setup}
    \end{flushleft}
    }]
}
\maketitle
\thispagestyle{empty}
\pagestyle{empty}

\begin{abstract}
Recent advances in data-driven reinforcement learning and motion tracking have substantially improved humanoid locomotion, yet critical practical challenges remain. In particular, while low-level motion tracking and trajectory-following controllers are mature, whole-body reference–guided methods are difficult to adapt to higher-level command interfaces and diverse task contexts: they require large, high-quality datasets, are brittle across speed and pose regimes, and are sensitive to robot-specific calibration. To address these limitations, we propose the Parameterized Motion Generator (PMG), a real-time motion generator grounded in an analysis of human motion structure that synthesizes reference trajectories using only a compact set of parameterized motion data together with high-dimensional control commands. Combined with an imitation-learning pipeline and an optimization-based sim-to-real motor parameter identification module, we validate the complete approach on our humanoid prototype ZERITH Z1 and show that, within a single integrated system, PMG produces natural, human-like locomotion, responds precisely to high-dimensional control inputs—including VR-based teleoperation—and enables efficient, verifiable sim-to-real transfer. Together, these results establish a practical, experimentally validated pathway toward natural and deployable humanoid control.

\end{abstract}

\section{INTRODUCTION}

Humanoid robotics continues to face three fundamental challenges: generating highly human-like motions, executing complex downstream tasks, and achieving reliable sim-to-real transfer. While prior work has addressed aspects of these challenges—ranging from single-segment motion imitation \cite{c17}, full-body teleoperation systems \cite{c21}\cite{c22}, and preliminary applications of human-like style in downstream tasks \cite{c18} —existing approaches are still difficult to adapt to higher-level command interfaces and diverse task contexts.

To overcome this limitation, we propose the \textbf{Parameterized Motion Generator (PMG)}, a unified framework for humanoid motion generation and control that enables natural human-like behavior, supports diverse command inputs and VR teleoperation, and facilitates effective sim-to-real transfer. PMG is grounded in a deep analysis of human motion patterns, requiring less than ten seconds of motion capture data to achieve parameterized modeling and synthesis of basic human-like motions. Its core mechanism constructs reference trajectories through dynamic interpolation and synthesis of parameterized motion units, while enforcing kinematic consistency via foot-contact constraints. This approach naturally supports data scalability, allowing adaptation to a wide range of motion styles. Furthermore, by employing an upper- and lower-limb separated motion tracking strategy, we integrate imitation learning with reinforcement learning to obtain robust motion policies in simulation.

    For experimental validation, we deployed the PMG algorithm on our self-developed humanoid robot, ZERITH Z1. It is worth noting that in the current research landscape, many successful deployments have been demonstrated on the Unitree G1, largely enabled by Unitree’s outstanding efforts in system identification and calibration, which have resulted in a remarkably small sim-to-real gap. However, such commercial products remain largely opaque to researchers due to proprietary internal optimizations. Since ZERITH Z1 is a prototype platform, we introduce an optimization-based sim-to-real parameter identification method for motor dynamic and static parameters, significantly reducing discrepancies between simulation and reality. As a result, PMG demonstrates, for the first time within a single system, human-like natural motion, precise responsiveness to high-dimensional control commands (velocity, orientation, and height) and VR operation, as well as efficient and verifiable sim-to-real transfer. In addition to its validated performance on ZERITH Z1, our PMG framework will be made publicly available to support reproducibility and future studies.

    The main contributions of this work are summarized as follows:
    
    
    
    
    \begin{itemize}

    \item \textbf{Parameterized motion generator:} We introduce a data-driven PMG that leverages a compact set of parameterized motion data to realize omnidirectional, full-pose, and human-like motion styles for versatile humanoid control.
    
    \item \textbf{Optimization-based humanoid SysID:} We develop a black-box optimization pipeline for motor parameter calibration, substantially narrowing the sim-to-real gap and enhancing policy deployability on hardware.
    
    \item \textbf{Real-world validation on prototype humanoid:} We validate the proposed framework on the non-commercial prototype ZERITH Z1, demonstrating natural locomotion and diverse task execution in real-world scenarios.
    
    \end{itemize}

\begin{figure*}[t]
    \centering
    \includegraphics[width=0.90\textwidth]{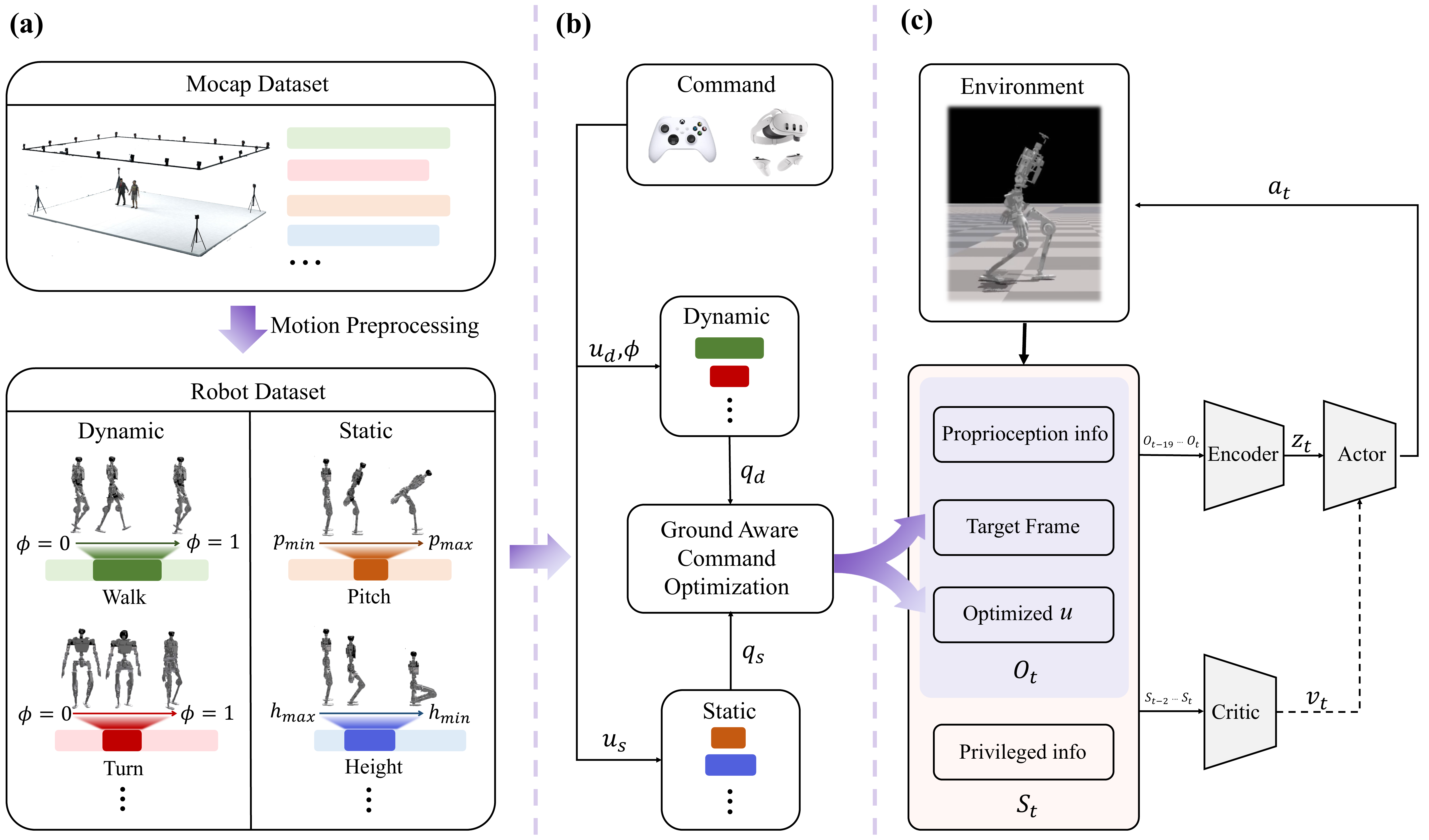}
    \caption{Overview of the Parameterized Motion Generator (PMG). The framework comprises three stages: 
    (a) \textbf{Motion Preprocessing} — raw mocap data (light bars) are retargeted to the robot and segmented into parameterized motion clips (darker bars). The database contains two types of clips: dynamic clips for locomotion, which are parameterized by phase $\phi$, and static clips for posture adjustment, which are parameterized by their admissible command ranges.
    (b) \textbf{Motion Generation} — high-level commands are decomposed into a dynamic (velocity) command $\mathbf{u}_d$ and a static (posture) command $\mathbf{u}_s$. The dynamic and static modules generate respective reference trajectories, which are subsequently refined by a Ground-Aware Command Optimization module. 
    (c) \textbf{RL Training / Deployment} — PMG outputs are incorporated into the observation and provided to the actor and critic; the actor's outputs are executed by low-level PD controllers and fed back to the environment.}
    
    \label{fig:framework}
    \vspace{-15pt}
\end{figure*}

\section{RELATED WORK}

\subsection{Humanoid Motion Generator}
Humanoid motion generators aim to produce natural and human-like motion sequences, demonstrating significant potential in robotic locomotion and motion control. Early works primarily relied on trajectory optimization or heuristic rules \cite{c1}, which are feasible in specific scenarios but struggle with highly dynamic tasks. Subsequently, in the fields of computer graphics and motion modeling, an increasing number of studies leverage real human motion data and deep generative models, such as VAEs, GANs and diffusion models, to generate conditional motions based on various inputs, including actions \cite{c5}\cite{c6}, text \cite{c7}\cite{c8}, or music \cite{c9}\cite{c10}. Some works have explored combining generative motion with robotic training, achieving partial success \cite{c11}\cite{c12}\cite{c13}\cite{c14}\cite{c15}. Nevertheless, generative models still face limitations, such as insufficient physical feasibility, limited real-time performance, heavy reliance on large-scale datasets and difficulty in precisely tracking commanded velocities and postures.      


\subsection{Human-like Locomotion}

To achieve human-like locomotion, traditional approaches have typically relied on carefully hand-crafted reward functions to induce human-like style. More recently, data-driven methods have become predominant: work in computer graphics \cite{c16} established the fragment-based imitation paradigm; in robotics, some studies achieve plausibly human-like behavior for individual motion primitives \cite{c17}\cite{c18}, while others leverage large-scale human motion datasets to construct full-body imitation or teleoperation systems \cite{c19}\cite{c20}\cite{c21}\cite{c22}. These methods centrally rely on precise tracking of keyframe or segment-level features and therefore depend heavily on high-quality, semantically consistent data; as a result, it is difficult to maintain both strong generalization and high tracking accuracy simultaneously. An alternative line of work adopts adversarial-style generation (e.g. \cite{c23} and its hardware or variant extensions), where a discriminator supplies style rewards and strict sequence-level matching is relaxed. Although adversarial approaches offer greater flexibility under weak supervision, they are often training-unstable on complex, multimodal datasets and prone to mode collapse.


\subsection{Sim-to-real Gap}
The \textit{sim-to-real} gap remains a critical challenge for humanoid robot applications, and several approaches have been proposed to mitigate this issue. \textbf{Domain Randomization (DR)} is a widely adopted strategy that introduces randomization into simulation environments and dynamic parameters (e.g., mass, friction coefficients, actuation delays) to enhance robustness against modeling errors \cite{c24}\cite{c25}\cite{c26}\cite{c27}. However, when the randomization range is excessively broad or heuristically designed, DR may fail to effectively narrow the sim-to-real gap and often results in overly conservative policies. \textbf{System Identification (SysID)} \cite{c28} provides an alternative approach by explicitly estimating or calibrating physical parameters to reduce discrepancies between simulation and reality. While SysID can improve policy transferability to real-world systems, both offline and online SysID remain challenging for high-dimensional and strongly nonlinear robotic systems. More recently, learning-based compensation methods (e.g. delta-action models \cite{c17} or residual learning) \cite{c29} have emerged, showing potential in improving sim-to-real transfer performance. Nevertheless, these methods often rely on black-box predictions, limiting their generalizability and interpretability.


\section{METHOD}
In this section, we elaborate on the implementation details of the PMG method. Our exposition is organized into three parts: efficient and accurate generation of human-like motions, effective network learning, and transfer to physical robots for downstream task execution.

\subsection{Parameterized Motion Generator}
\subsubsection{Motion Preprocessing}

As illustrated in Figure \ref{fig:framework}, PMG relies on high-quality motion data. We first collect two categories of human motions using a high-precision motion capture system: dynamic motions (e.g., walking, turning), denoted as $D_{\text{human}}^{d}$, and static postures (e.g., squatting, bending), denoted as $D_{\text{human}}^{s}$. Each motion is designed as a single-directional displacement or posture adjustment, enabling motion decoupling for subsequent decomposition and recomposition within PMG.

Subsequently, the collected data are retargeted to our in-house humanoid robot ZERITH Z1 via optimization-based algorithms \cite{c19} \cite{c30}, resulting in two robot datasets:

\begin{itemize}
    \item Dynamic dataset:
    \begin{equation}
        D_{\text{robot}}^{d} = \{q, \dot{q}, v, \omega, c\}
    \end{equation}
    where $q$ denotes joint angles, $\dot{q}$ joint angular velocities, $v$ base linear velocity, $\omega$ base angular velocity, and $c = \{\mu, \sigma\}$ denotes foot contact information, including contact center and contact range.
    
    \item Static dataset:
    \begin{equation}
        D_{\text{robot}}^{s} = \{q, p, r, h\}
    \end{equation}
    where $p$ denotes pitch, $r$ roll, and $h$ the base height.
\end{itemize}

Since $D_{\text{robot}}^{d}$ only contains primitive motion segments, it exhibits periodicity. We align and clip each segment using the contact information $c$ to preserve exactly one gait cycle. A Gaussian smoothing is then applied at the boundary to avoid discontinuities. The final preprocessed dataset is expressed as:
\begin{equation}
    D_{\text{robot}}^{\text{d-clip}} = \{q, \dot{q}, v, \omega, c, T\}
\end{equation}
where $T$ represents the duration of one gait cycle.

\subsubsection{Parameterized Gait Generation}
\label{sec:parameterized_gait_generation}
Given a command consisting of velocity $\mathbf{u}_d=[v_x, v_y, \omega]$ and posture $\mathbf{u}_s=(p,r,h)$, together with a gait phase $\phi \in \left[0,1\right)$, PMG synthesizes reference joint trajectories through a unified weighted interpolation framework. 

For each velocity channel $x\in\{v_x, v_y,\omega\}$, we retrieve a phase-dependent motion template from the dataset and interpolate it with the standing posture $q_{\mathrm{stand}}$ according to the normalized magnitude factor:
\begin{equation}
\alpha_x = \mathrm{Clip}\!\left(\frac{|u_{d,x}|}{\bar{x}},\,0,\,1\right)
\end{equation}
where $\bar{x}$ is the nominal scale of channel $x$. The channel-wise interpolated motion is then:
\begin{equation}
\tilde{q}_x(\phi) = \alpha_x \,q_{D_x}(\phi) + (1-\alpha_x)\,q_{\mathrm{stand}}
\end{equation}

To ensure non-negativity and convexity of the mixture, we define normalized weights:
\begin{equation}
w^{(x)} = \frac{|u_{d,x}|}{\sum_{i\in\{x,y,\omega\}} |u_{d,i}| + \varepsilon}
\end{equation}
where $\varepsilon>0$ is a small constant for numerical stability. The dynamic reference is obtained as a convex combination, let \(\mathcal I=\{x,y,\omega\}\):

\begin{equation}
q_{d}(\phi) \;=\; \sum_{i\in\mathcal I} w^{(i)}\,\tilde{q}_i(\phi)
\end{equation}

Contact-related variables are interpolated analogously:
\begin{align}
r_{d} &= \sum_{i\in\mathcal I} w^{(i)}\, r_i,\\
\mu_{d} &= \sum_{i\in\mathcal I} w^{(i)}\, \mu_i.
\end{align}

Similarly, posture commands $(p,r,h)$ are normalized into coefficients $\beta_{\cdot}\in[0,1]$ and interpolated with their respective templates to yield a posture offset $q^{s}$. The final joint reference is given by:
\begin{equation}
q_{\mathrm{ref}}(\phi) = q_{d}(\phi) + q_{s}
\end{equation}

At each time step, $\phi$ is incremented by a normalized term that depends on the weighted average period $T_u$: 
\begin{equation}
\phi_{\mathrm{new}} = \left(\phi + \frac{\Delta t}{T_u}\right) \bmod 1
\end{equation}
where $T_u$ is computed as the weighted average of the candidate periods:

\begin{equation}
T_u = \sum_{i \in \mathcal I} w^{(i)}\, T_i,
\end{equation}

The reference joint velocity is obtained by discretizing the temporal difference of the reference joint trajectory.

This convex combination framework guarantees that the synthesized motion remains within the feasible span of the dataset templates while preserving smooth dependence on the command input. However, since interpolation does not strictly enforce kinematic consistency with the commanded velocities, discrepancies such as foot slippage may occur, necessitating further ground-aware optimization.
\begin{figure}[h]
    \centering
    \includegraphics[width=\columnwidth]{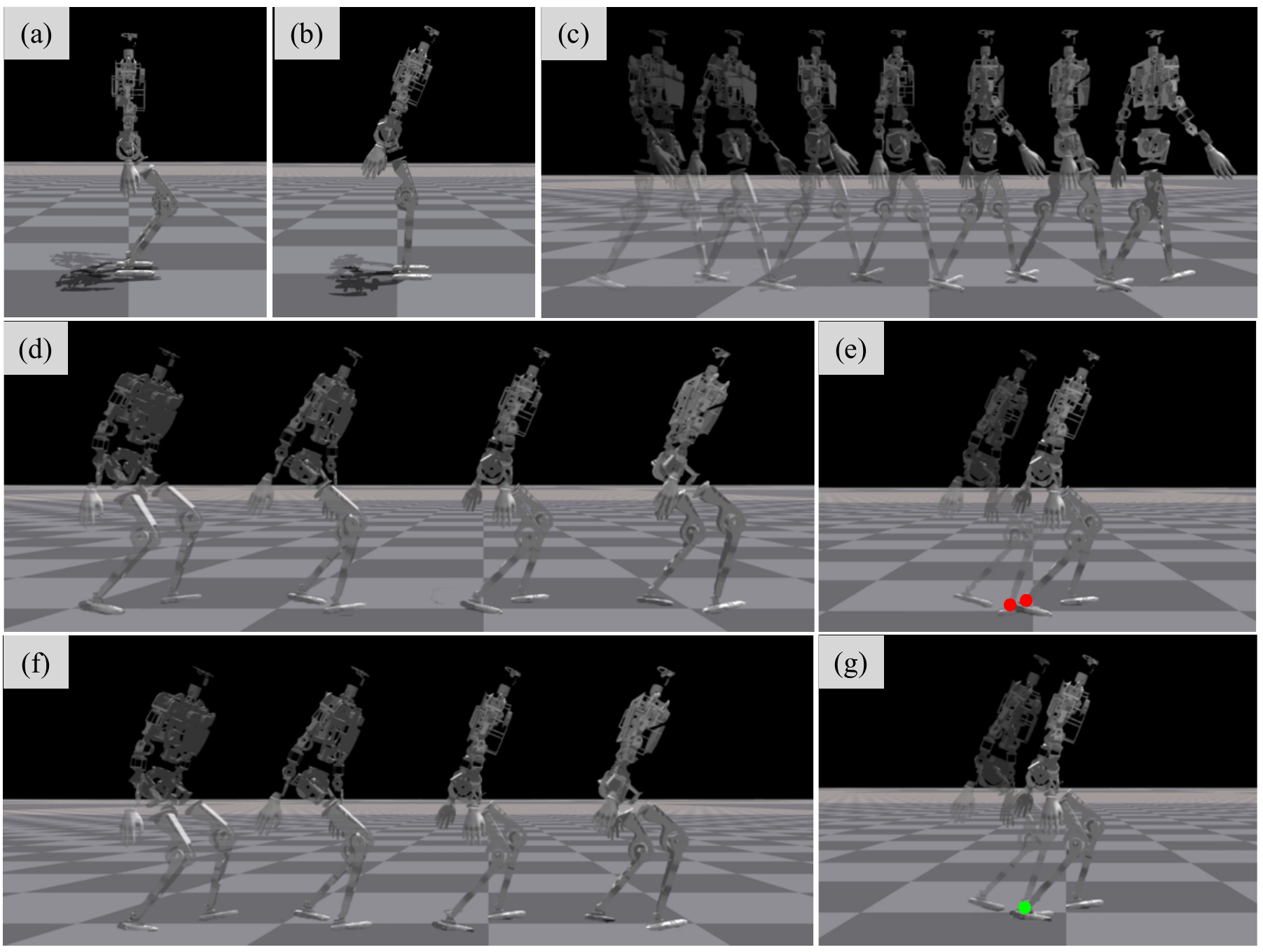}
    \caption{PMG motion comparison. All sequences are open-loop reference motions and have not been fine-tuned with reinforcement learning. 
    (a–c) Reference clips from the robot dataset: (a) squatting from $D_{\mathrm{robot}}^{s}$, (b) forward-bend (pitch) from $D_{\mathrm{robot}}^{s}$, and (c) forward locomotion from $D_{\mathrm{robot}}^{d}$. (d–g) Synthesized motions: (d) synthesized motion without Ground-Aware Command Optimization (GCO); (e) the same sequence with red markers highlighting pronounced foot slip (physically inconsistent contact); (f) synthesized motion after applying GCO; (g) overlay showing that foot slip is substantially corrected.}

    \label{fig:example}
    \vspace{-15pt}
\end{figure}
\subsubsection{Ground-aware Command Optimization}
As discussed in Section~\ref{sec:parameterized_gait_generation}, the reference motions synthesized via weighted template interpolation do not always match the input commands. For instance, in certain situations, the commanded base height and pitch may conflict, or the maximum linear and angular velocities cannot be realized simultaneously due to gait constraints. To address this, we adopt a ground-aware optimization strategy that enforces consistency between the generated motion and the commanded inputs.

Specifically, given the foot contact information $c_f(\phi)\in\{0,1\}$ for each foot $f\in\{L,R\}$, we require that feet predicted to be in contact with the ground remain stationary. Using forward kinematics, the foot positions and velocities are computed from the base pose $\mathbf{x}_{\mathrm{base}}(\phi)$ and velocity $\mathbf{v}_{\mathrm{base}}(\phi)$:
\begin{equation}
\mathbf{p}_{\mathrm{foot},f}(\phi) = \mathbf{x}_{\mathrm{base}}(\phi) + \mathbf{p}^{\mathrm{leg}}_f(q_{\mathrm{leg},f}(\phi))
\end{equation}

\begin{equation}
\quad
\mathbf{v}_{\mathrm{foot},f}(\phi) = \mathbf{v}_{\mathrm{base}}(\phi) + \dot{\mathbf{p}}^{\mathrm{leg}}_f(q{_\mathrm{leg},f},\dot{q}_{\mathrm{leg},f})
\end{equation}
where $\mathbf{p}^{\mathrm{leg}}_f(\cdot)$ and $\dot{\mathbf{p}}^{\mathrm{leg}}_f(\cdot)$ denote the foot position and velocity obtained from the leg forward kinematics.

For each foot in contact ($c_f(\phi)=1$), we enforce the following static constraint:

\begin{equation}
\mathbf{v}_{\mathrm{foot},f}(\phi), \omega^{z}_{\mathrm{foot},f}(\phi)=\mathbf{0}, \quad (\mathbf{p}_{\mathrm{foot},f}(\phi))_z = h_{\mathrm{ground}}
\end{equation}

and compensate the base velocity and height accordingly:
\begin{equation}
\mathbf{v}'(\phi) = \mathbf{v}_{\mathrm{base}}(\phi) + \Delta \mathbf{v}_{\mathrm{base}}
\end{equation}
\begin{equation}
\quad
\omega'(\phi) = \omega_{\mathrm{base}}(\phi) + \Delta \omega_{\mathrm{base}}
\end{equation}
\begin{equation}
\quad
h'(\phi) = h_{\mathrm{base}}(\phi) + \Delta h_{\mathrm{base}}
\end{equation}

The optimized command vector is then:
\begin{equation}
\mathbf{u}'(\phi) = [\mathbf{v}'(\phi),\ \omega'(\phi),\ \mathrm{pitch}(\phi),\ \mathrm{roll}(\phi),\ h'(\phi)]
\end{equation}

This approach effectively eliminates base drift and foot suspension artifacts caused by naive template interpolation, ensuring kinematic consistency with ground-contact constraints. However, since only kinematic (static) constraints are considered, dynamic feasibility is not guaranteed. Therefore, reinforcement learning or a dynamics-aware optimization is still required to ensure stability during dynamic execution.

\subsection{Network Training}

We employ an \emph{asymmetric actor-critic} network for imitation learning. To improve task generalization and facilitate diverse applications, we adopt a \emph{limb-separated training} scheme: the actor outputs only lower-body joint actions, while the upper-body joints follow open-loop reference trajectories generated by PMG. During training, a subset of upper-body joint actions are perturbed with smooth, continuous random values in joint space to enhance robustness and teleoperation adaptability.

The reward design follows standard motion-tracking frameworks and consists of two main components: a task reward and an energy penalty. The task reward is designed to encourage the robot to follow the commanded velocities and postures, while also tracking reference joint positions, joint velocities, and foot-ground contact patterns. The energy penalty discourages excessive torque or energy consumption. 

\subsection{Sim-to-Real Transfer}



\subsubsection{Motor dynamic performance calibration}

We observed substantial discrepancies between real actuators and the idealized motor model used in simulation, which degrade policy transfer. To correct this at the actuator level, we perform single-joint response calibration: while holding other joints fixed, we apply a diverse excitation trajectory \(q_{\mathrm{cmd}}(t)\) (mixture of steps and smooth segments) to a single motor and record the real response \(q_{\mathrm{real}}(t)\) for \(t=1,\dots,T\). In simulation we reproduce the same excitation and denote the simulated response by \(q_{\mathrm{sim}}(t;\eta)\), where the simulator parameter vector is chosen as \(\eta=[K_p,\;K_d,\;I]^\top\) (proportional and derivative gains and equivalent rotor inertia; additional terms such as damping or friction may be appended if required). We seek \(\eta\) that minimizes the trajectory mismatch; a practical alignment loss is:
\begin{equation}
    L(\eta)=\sum_{t=1}^{T} \big\|q_{\mathrm{sim}}(t;\eta)-q_{\mathrm{real}}(t)\big\|^2
\end{equation}

The identification problem is \(\eta^\star=\arg\min_{\eta}L(\eta)\). We solve this black-box optimization by a sampling-based optimizer CMA-ES: at each iteration a population \(\{\eta^{(i)}\}_{i=1}^M\) is sampled, \(L(\eta^{(i)})\) is evaluated in parallel across Isaac Gym instances, and the sampling distribution is updated toward low-loss regions until convergence. Subsequently, we used the joint parameters obtained through optimization for policy training.

\subsubsection{Motor zero-point calibration}

After completing the motor dynamic calibration, we observed that structural deformation of the motor mounting parts introduces significant zero-point offsets. Due to the lack of external measurement devices, we propose a calibration method based on Motor–IMU alignment. The key idea is to enforce consistency between the IMU readings of the real robot and the simulated robot under multiple default poses, and to use the discrepancy in joint angles as zero-point offsets for updating the motor firmware.

Formally, for joint \(i\) at pose \(k\), let the simulated joint angle be \(q^{\mathrm{sim}}_{i,k}\), the raw 
encoder reading be \(r_{i,k}\), and the current zero-point estimate be \(z_i\). The real joint angle estimate is then:

\begin{equation}
    q^{\mathrm{real}}_{i,k} = r_{i,k} + z_i
\end{equation}

We only consider samples where IMU alignment is achieved:
\begin{equation}
    \|\mathbf{s}^{\mathrm{real}}_{k} - \mathbf{s}^{\mathrm{sim}}_{k}\| \le \tau
\end{equation}
where \(\mathbf{s}\) denotes the IMU signal and \(\tau\) is a predefined threshold. The residual zero-point error is computed as:
\begin{equation}
    \delta_{i,k} = q^{\mathrm{sim}}_{i,k} - q^{\mathrm{real}}_{i,k}
\end{equation}

Aggregating across multiple poses yields the correction term  by median:
\begin{equation}
    \delta_i = \mathrm{median}_k\{\delta_{i,k}\}
\end{equation}

The zero-point is iteratively updated with damping:
\begin{equation}
    z_i \leftarrow z_i + \alpha\,\delta_i
\end{equation}
where \(\alpha\in(0,1]\) is a damping factor. Convergence is declared if \(|\alpha\,\delta_i| < \varepsilon\) for all joints over \(M\) consecutive iterations.

This method eliminates the need for external measurement equipment, automatically estimating and correcting zero-point offsets through Motor–IMU alignment. Empirical evaluation shows that it can compensate for offsets larger than \(0.02\) rad, effectively mitigating systematic errors caused by structural deformation.

\subsubsection{Domain randomization}

To further bridge the sim-to-real gap, we apply domain randomization to hardware-related and sensor-related parameters. We design a series of calibration experiments to empirically determine realistic parameter ranges, ensuring that the randomized distributions remain physically meaningful. Excessively large ranges may lead to overly conservative policies; thus, we carefully balance robustness and performance. 




















\begin{figure*}[t]
    \centering
    \includegraphics[width=0.94\textwidth]{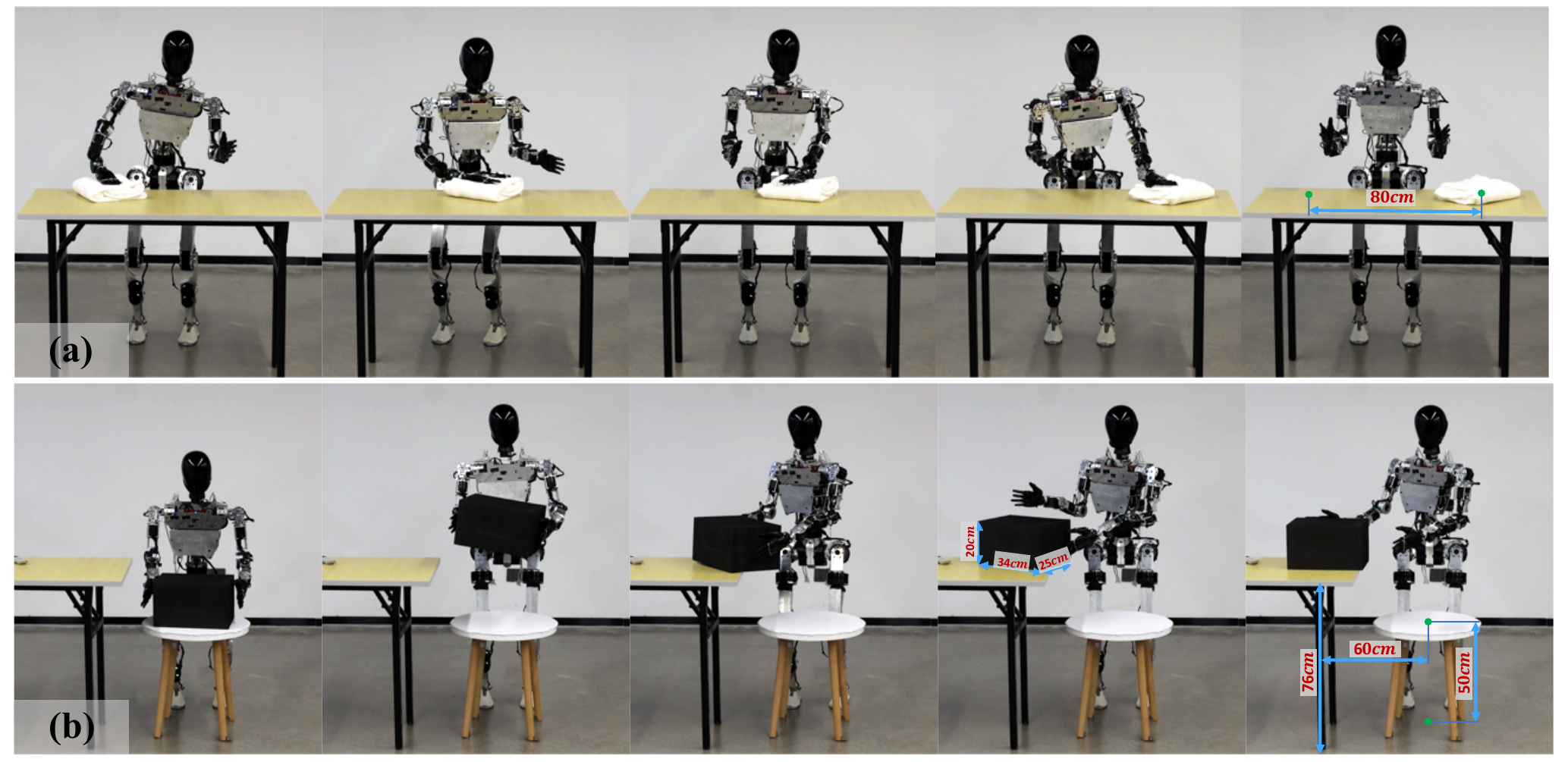}
    \caption{Real-world teleoperation experiments with the humanoid robot ZERITH Z1.
(a) Wiping task: the robot moves a towel laterally across the tabletop for over $80 cm$.
(b) Box-picking task: The robot lifts a black box (width $34\,\mathrm{cm}$) from a table of height $50\,\mathrm{cm}$, 
translates it horizontally by $\approx 60\,\mathrm{cm}$ to another table of height $76\,\mathrm{cm}$, 
and places it down.}
    \label{telep_exp}
\end{figure*}

\section{EXPERIMENT}

This section presents a systematic evaluation of the proposed Parameterized Motion Generator (PMG), focusing on both performance and effectiveness. The experiments consist of two parts: (1) comparative and ablation studies in simulation, which validate the algorithm design and assess the contribution of each component. (2) real-world experiments on our humanoid robot ZERITH Z1, including locomotion and teleoperation tasks, which demonstrate sim-to-real transfer and applicability to downstream tasks. To ensure fair comparison, all policies are trained with identical reward functions, hyperparameters, and domain randomization settings.

\subsection{Experiment Setting}
\subsubsection{Robot and Hardware Platform}

We employ the self-developed humanoid robot prototype ZERITH Z1 as the experimental platform. The robot stands $150$ cm tall, weighs $35$ kg, and has a total of $31$ degrees of freedom (DoFs): $6$ DoFs per leg, $3$ DoFs in the waist, $7$ DoFs per arm, and $2$ DoFs in the head (which are not utilized in this work). Low-level motor control is performed via an EtherCAT–CAN bus at a frequency of $500$ Hz. High-level policy inference is executed on a low-cost x86 host (UP710s), optimized with C++ and the ONNX runtime, enabling stable neural network inference and PMG control at $100$ Hz.

For interaction, a commercially available game controller is used to command robot movement and posture adjustments via its joysticks and buttons. In the teleoperation experiments, a VR device (Meta Quest $3$) is employed: the headset's position and orientation control the robot’s global movement and posture, while the relative pose of the controllers with respect to the headset is interpreted through inverse kinematics to compute joint angles, thereby enabling remote teleoperation of the robot’s arms.






\subsubsection{Training Setup}

All policies are trained using the PPO algorithm with Adam optimizer, and parallelized environment rollouts. Training is performed on an NVIDIA RTX $4090$ GPU. Under this setup, policies converge within approximately 8 hours.

\begin{table}[h]
\caption{Command Tracking Error}
\label{AblationExperiment}
\renewcommand{\arraystretch}{1.3}
\centering 
\begin{tabular}{l c c c c c }

\toprule
Method& LinVel& YawVel& Roll& Pitch& Height\\
\midrule
Baseline&  \textbf{0.0824}&  0.4121& 0.1966& 0.3891& 0.0873     \\
PMG w/o GCO& 0.3762& 0.5131&  0.1595& 0.1981&  0.0656\\
PMG (Ours)& 0.1228& \textbf{0.1612}& \textbf{0.0769}& \textbf{0.0697} & \textbf{0.0351}    \\
\bottomrule

\end{tabular}


\end{table}

\subsection{Simulation Ablation Studies}
\subsubsection{Experimental Configurations}

To analyze the contribution of individual components, we compare the following three configurations:

\textbf{Single Gait Imitation Policy}: This configuration directly imitates a single reference walking gait, without leveraging the Parameterized Motion Generator (PMG) introduced in this work. It serves as an upper-bound baseline for performance under pure imitation learning.


\textbf{PMG w/o GCO}: This configuration employs the PMG framework proposed in this work, but excludes the Ground-aware Command Optimization (GCO) module.


\textbf {PMG (Ours)}: The full PMG framework, including all modules described above.

\subsubsection{Evaluation Procedure} All three policies are evaluated within the Isaac Gym simulator under an identical sequence of preset task commands. During execution, no external intervention is applied even in cases of severe gait deformation or falling. We employ the following quantitative metrics for evaluation:


    

\begin{itemize}
    \item \textbf{Command Tracking Error}: Measured by the errors in linear velocity, angular velocity, torso pitch, torso roll, and body height.
\end{itemize}

\subsubsection{Results and Discussion}


As shown in Table \ref{AblationExperiment}, \textbf{Single Gait Imitation} performs well only along velocity directions that match its reference gait and degrades markedly under other commands. \textbf{PMG w/o GCO} sacrifices velocity-tracking accuracy but yields noticeably better posture-tracking than the baseline. The full \textbf{PMG} matches the baseline on linear-velocity tracking and outperforms both alternatives on all other commands, validating PMG for omnidirectional motion and posture regulation and underscoring the necessity of the Ground-aware Command Optimization (GCO) module.

\begin{figure}[t]
    \centering
    \includegraphics[width=\columnwidth]{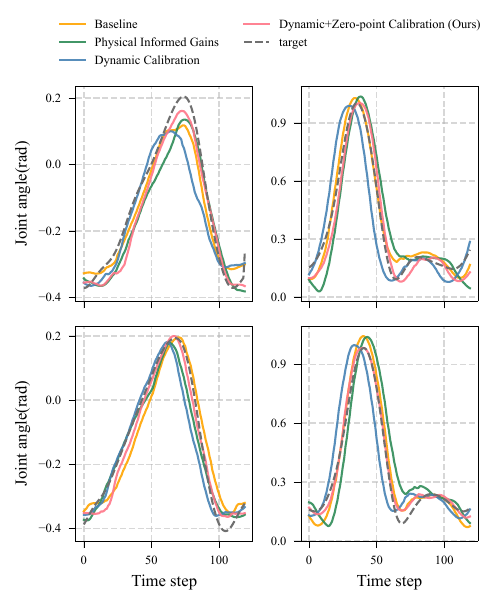}
    \caption{Tracking Curves Between Actual Joint Angles and Target Joint Angles Generated by Gait Generator.}
    \label{fig:real_world_experiment}
\end{figure}


\begin{table}[h]
\caption{Sim-to-Real Evaluation}
\label{realworldExperiment}
\renewcommand{\arraystretch}{1.3}
\centering 
\resizebox{\linewidth}{!}{%
\begin{tabular}{l c c}
\toprule
Policies & Motion Imitation Error & Sim-to-Real Error\\
\midrule
Baseline & 0.0571 & 0.0781 \\
Physically Informed Gains & 0.0747 & 0.0656\\
Dynamic Calibration & 0.0682 & 0.0562\\ 
Ours & \textbf{0.0562} & \textbf{0.0414}\\
\bottomrule
\end{tabular}%
}
\end{table}


\subsection{Sim-to-Real Evaluation}
\subsubsection{Experimental Configurations}
To evaluate the generalization ability and sim-to-real transfer performance of the proposed framework, we trained and deployed four policies under identical reward functions:

\textbf{Baseline}:
Policies are trained and deployed using a set of PD parameters that perform reasonably well in simulation, without any adaptation to the hardware.

\textbf{Physically Informed Gains}: 
PD parameters are configured based on motor rotor inertia and other physical properties, following the computation formulas reported in \cite{c18}, aiming to make joint responses more consistent with real hardware dynamics.

\textbf{Dynamic Calibration}: 
The hardware uses the default PD parameters, while applying motor dynamic performance calibration as described in this work.

\textbf{Dynamic+Zero-point Calibration (Ours)}: Using both motor dynamic calibration and motor zero-point calibration.

\subsubsection{Evaluation Procedure}

 In this experiment, we primarily focus on sim-to-real performance, using a forward locomotion task as a representative test scenario. All policies are initialized from the same standing pose. After 5 seconds of standing, a preset sequence of continuous forward velocity commands is issued. We record joint positions, policy actions, and the target joint angles generated by the PMG in both simulation and hardware experiments. 


\begin{itemize}
    \item \textbf{Sim-to-Real Error}: We compare the joint position errors in simulation and on the real robot under identical environment settings and command inputs to characterize the sim-to-real error. To ensure accuracy, multiple seeds are used in simulation, and repeated trials are conducted on the hardware.
    
    \item \textbf{Motion Imitation Error}: Given that PMG deterministically generates the same motion sequence under identical commands, we compare how different policies track the same reference motion on hardware, in order to assess which policy achieves superior realization of PMG’s tracking capability. 
\end{itemize}

\subsubsection{Results and Discussion}
As shown in Figure \ref{fig:real_world_experiment} and Table \ref{realworldExperiment}, our method achieves the lowest error in both Motion Imitation Error and Sim-to-Real Error, demonstrating its effectiveness in reducing the sim-to-real gap and enhancing the performance of imitation learning policies.

\subsection{Teleoperation task Evaluation}
\subsubsection{Experimental Configurations}  To evaluate the capability of our framework in full-body control and downstream task applications, we design two teleoperation tasks that require coordinated upper-limb control and whole-body postural regulation: a wiping task and a box-picking task.

\subsubsection{Evaluation Procedure}
In real-world experiments, each task is executed for $20$ trials. To increase test variability, certain task initial conditions are randomized: for the wiping task the towel is placed at a random position along one side of the table (within the reachable region), and for the box-picking task the box's initial position and orientation are randomized. All other environmental and control settings remain identical across trials. We employ the following quantitative metric:

\begin{itemize}
    \item \textbf{Success rate:} Denote by $N_{\mathrm{succ}}$ the number of successful trials out of $20$. The success rate is computed as
    \[
      \mathrm{SR} = \frac{N_{\mathrm{succ}}}{20}.
    \]
    A trial is considered successful if it meets the task-specific criterion: for the wiping task, the towel is moved laterally across the tabletop for a distance exceeding $1\,\mathrm{m}$; for the box-pickup task, the robot lifts the box to approximately $30\,\mathrm{cm}$, translates it horizontally by about $60\,\mathrm{cm}$, and places it down smoothly without dropping. 
\end{itemize}


\begin{table}[h]
\caption{Teleoperation task Evaluation}
\label{Teleoperation}
\renewcommand{\arraystretch}{1.3}
\centering 
\begin{tabular}{l c c}

\toprule
  & Wiping task & Box-picking task\\
\midrule
Success Rate& 0.95 &  0.8 \\

\bottomrule

\end{tabular}


\end{table}

\subsubsection{Results and Discussion} As shown in Figure \ref{telep_exp} and Table \ref{Teleoperation}, the robot is able to successfully complete most teleoperation tasks, achieving an overall high success rate. However, due to the unavailability of dexterous hands and the absence of haptic feedback, failures still occur in certain cases (e.g., collisions or loss of balance).





\section{CONCLUSIONS}

In this work, we propose a Parameterized Motion Generator (PMG) for humanoid robots and validate it on our in-house platform, ZERITH Z1, through a systematic sim-to-real pipeline. Experiments are conducted on both locomotion and teleoperation tasks. The results demonstrate that PMG can generate natural and human-like reference motions, exhibiting smooth gait transitions and large-range whole-body posture variations.

Nevertheless, as a primarily data-driven approach, PMG remains heavily dependent on high-quality reference data and is thus more suited for periodic, walking-related behaviors, while its ability to handle complex, non-periodic motions such as dancing is still restricted. To address these challenges, future work will focus on incorporating richer dynamics constraints, improving the integration of imitation learning and RL, and expanding the diversity of reference datasets to enhance generalization and robustness.











\end{document}